\documentclass{article}

\usepackage[preprint]{corl_2024} 

\usepackage{amsmath}
\usepackage{booktabs}
\usepackage{graphicx}
\usepackage{subcaption}
\usepackage{tabularx}
\usepackage{xcolor}
\usepackage{xspace}

\newcommand{\method}{\textsc{RoboPoint}\xspace}
\newcommand{\dataset}{\textsc{Where2Place}\xspace}

\title{\includegraphics[height=9mm]{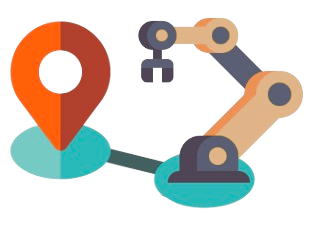}RoboPoint: A Vision-Language Model for Spatial Affordance Prediction for Robotics}

\author{
  Wentao Yuan$^1$ \And Jiafei Duan$^1$ \And Valts Blukis$^2$ \And Wilbert Pumacay$^4$ \And Ranjay Krishna$^{1,3}$
  \And Adithyavairavan Murali$^2$ \And Arsalan Mousavian$^2$ \And Dieter Fox$^{1,2}$ \Inst
  $^1$University of Washington \Inst $^2$NVIDIA \INST $^3$Allen Institute for Artifical Intelligence \Inst $^4$Universidad Católica San Pablo
}

\begin{document}
\maketitle
\vspace{-15pt}

\begin{abstract}
From rearranging objects on a table to putting groceries into shelves, robots must plan precise action points 
to perform tasks accurately and reliably.
In spite of the recent adoption of vision language models (VLMs) to control robot behavior, 
VLMs struggle to precisely articulate robot actions using language. 
We introduce an automatic synthetic data generation pipeline that instruction-tunes VLMs to robotic domains and needs. 
Using the pipeline, we train \method\ , a VLM that predicts image keypoint affordances given language instructions. 
Compared to alternative approaches, our method requires no real-world data collection or human demonstration, making it much more scalable to diverse environments and viewpoints.
In addition, \method\ is a general model that enables several downstream applications such as robot navigation, manipulation, and augmented reality (AR) assistance. Our experiments demonstrate that \method\ outperforms state-of-the-art VLMs (GPT-4o) and visual prompting techniques (PIVOT) by 21.8\% in the accuracy of predicting spatial affordance and by 30.5\% in the success rate of downstream tasks.
Project website: \href{https://robo-point.github.io/}{robo-point.github.io}.
\end{abstract}

\keywords{Foundation Model, Affordance Prediction, Open-world Manipulation} 

\section{Introduction}
\label{sec:intro}

Spatial reasoning is fundamental to all intellectual processes~\cite{tversky2009thinking}. Beyond its prominence in understanding geometry, science, and architecture~\cite{taylor1992spatial}, spatial reasoning significantly impacts our everyday lives. Even mundane tasks like purchasing groceries require us to identify the vacant space in our shopping carts to load more items. One critical mechanism through which we communicate plans that involve navigation and manipulation is by \emph{pointing}. Studies in developmental psychology demonstrate that infants and adults alike point to share information about their environment~\cite{tomasello2007new}. In robotics, pointing has been operationalized through waypoints for navigation and task execution. Roboticists have found that when robots use waypoints effectively, it mimics human pointing, leading to more intuitive plans~\cite{dragan2013legibility}.

Recent explorations have cast aside pointing in favor of language instructions with the advent of large VLMs~\cite{achiam2023gpt,bai2023qwen,liu2023llava}. Trained on large datasets of images and language, VLMs can provide powerful visual semantic understanding and useful guidance to robotic tasks, such as which object a manipulator should pick up or which goal a mobile robot should reach~\cite{brohan2023can,liang2023code,singh2023progprompt}. However, language is not precise enough to successfully guide robot behavior. Even the most recent and powerful VLMs, such as GPT-4o~\cite{openai2024gpt4o}, have limited accuracy in real robot execution, especially when language commands use spatial relations to identify objects or refer to object-free locations, such as ``place the cup next to the plate".

In this work, we introduce \method, an open-source VLM instruction-tuned to \emph{point}. Two key features differentiate \method from other VLMs for robotics: a \textbf{point-based action space} and a \textbf{scalable data pipeline}. First, inspired by prior works~\cite{mo2021where2act,shridhar2022perceiver,goyal2023rvt}, we fine-tune \method using \emph{spatial affordance prediction}, the task of pointing at where to act. The actions are specified via points in the RGB image, and then transformed to 3D using depth information, removing the need for pre-defined action primitives~\cite{liang2023code,singh2023progprompt}, external object detectors~\cite{huang2023voxposer,liu2024moka}, or iterative visual prompting~\cite{nasiriany2024pivot}.
Second, we design a fully autonomous pipeline generating a large, diverse dataset of ground truth action points by computing spatial relations from the camera's perspective and sampling points within object masks and object-surface intersections. Compared to approaches that require expensive human demonstration data~\cite{brohan2022rt,brohan2023rt,team2024octo}, our pipeline is much easier to scale. Even though we only added data containing simulated images along with templated language,  the resulting model's performance improves on real images with natural language commands.

Our results show that \method\ significant outperforms various powerful VLMs such as GPT-4o~\cite{openai2024gpt4o}, LLaVA-NeXT~\cite{liu2024llavanext}, Qwen-VL~\cite{bai2023qwen} and SpatialVLM~\cite{chen2024spatialvlm} on relational object reference, free space reference and object rearrangement in cluttered, real-world environments, without losing accuracy on standard VQA benchmarks. To evaluate relational free space reference, we collect \dataset, a manually annotated, challenging real-world benchmark. We also show very promising results beyond robotic applications in an interactive augmented reality (AR) setting, where \method provides visual action suggestions, effectively guiding users through tasks by predicting target points based on common sense.


\begin{figure}
    \centering
    \vspace{-2em}
    \includegraphics[width=0.85\linewidth]{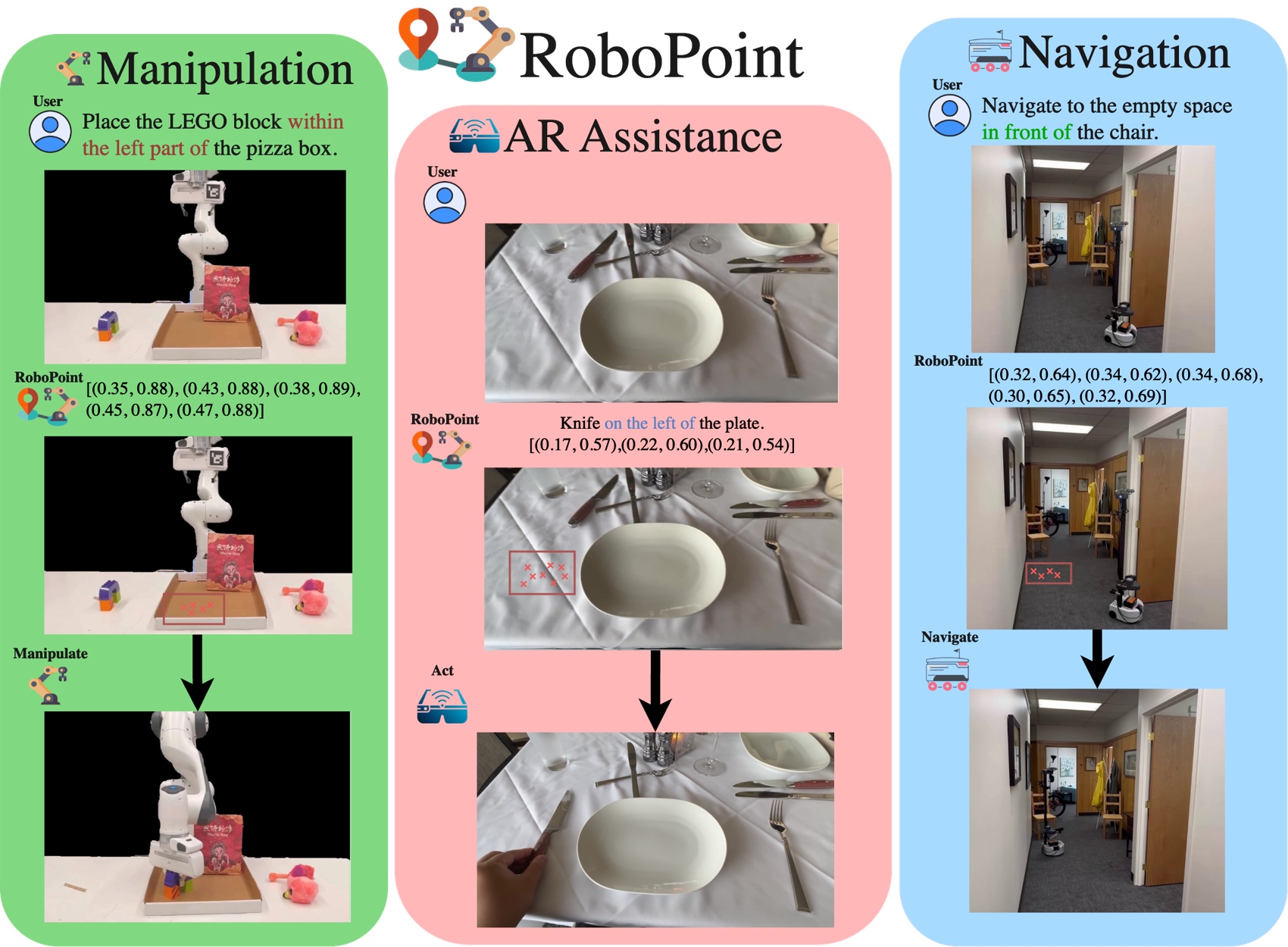}
    \caption{\small \method is a \textbf{Vision-Language Model} that predicts \textbf{affordance points} based on language instructions. It is able to generate precise actions (red crosses in the image) which satisfy spatial relations in the instruction. \method\ is a generic VLM that can be applied to many domains such as manipulation, augmented reality and navigation.}
    \vspace{-15pt}
    \label{fig:teaser}
\end{figure}

\section{Related Work}
\label{sec:related}
Inspired by prior works on \textit{spatial reasoning} and \textit{affordance prediction}, \method\ takes a distinct approach to build VLMs for robotics in contrast to recent methods using \textit{zero-shot language models}.

\paragraph{Spatial Reasoning} 
Many VQA benchmarks \cite{yu2016modeling,krishna2017visual,hudson2019gqa,yi2019clevrer,duan2022survey,duan2023ar2} have included problems about spatial relations as indicator for a model's ability to understand 3D. These problems can be solved using state estimation \cite{xiang2017posecnn} plus symbolic reasoning \cite{ding2020object}, but these methods have poor generalization to novel objects. More recently, SORNet \cite{yuan2022sornet} shows that a transformer model conditioned on object prompts can generalize zero-shot to unseen objects on spatial reasoning tasks, similar in spirit to modern VLMs. However, existing works on spatial reasoning mostly focused on coarse-grained relations. SpatialVLM \cite{chen2024spatialvlm} took a step forward to predict spatial relations in metric space, but we show that \method\ can achieve better performance on real-world spatial reasoning tasks by locating affordances as points.

\paragraph{Affordance Prediction}
Affordance is defined as the functions of a object, i.e. in what ways it can be manipulated. It goes beyond the visual properties and ties observations to actions. The efficacy of affordance prediction has been shown by many learning-based manipulation methods for 6-DoF grasping \cite{sundermeyer2021contact,murali2021same,jiang2021synergies} and stable object placement \cite{zeng2020transporter,liu2022structformer,yuan2023m2t2}. Affordance can be represented in many ways such as part segmentation \cite{do2018affordancenet}, dense image feature descriptors \cite{florencemanuelli2018dense} and keypoints \cite{manuelli2019kpam,qin2020keto,mo2021where2act}. We use the 2D keypoint representation to train \method\ since it can be readily converted into language format.

\paragraph{Zero-shot Language Models for Robotics}
Several works \cite{brohan2023can,liang2023code,singh2023progprompt} have shown that language model are capable planners for robotic tasks. Using in-context learning \cite{brown2020language}, these methods generate reasonable plans in structured language, but requires pre-defined action primitives to execute. More recent works leverage VLMs to generate more fine-grained outputs. VoxPoser \cite{huang2023voxposer} generates 3D value maps. PIVOT \cite{nasiriany2024pivot} iteratively samples and evaluates possible actions in image space. MOKA \cite{liu2024moka} predicts keypoints specific to an action type. Unlike \method, all of these approaches still rely on external models for detecting objects relevant for the task.

\section{Method}
\label{sec:method}
\method\ is instruction-tuned from Vicuna-v1.5-13B \cite{vicuna2023} with a mix of synthetic and real-world data on spatial affordance prediction. This section will cover 3 critical aspects of the tuning pipeline: 1) the problem formulation 2) the instruction tuning procedure and 3) the curation of the data mix.

\paragraph{Spatial Affordance Prediction}
We formulate the problem of spatial affordance prediction as predicting a set of target point coordinates $\{(x_0, y_0), (x_1, y_1), ..., (x_n, y_n)\}$ in image space that satisfy the relations indicated by a language prompt. This formulation has several advantages. First, compared to fuzzy language actions such as ``place the apple in the drawer", which requires detection of apple and drawer before execution, a point prediction is much more precise and can be directly converted to actions. Most VLMs are trained to predict bounding boxes. However, from Fig.~\ref{fig:qualitative}, we can see that bounding boxes often include a lot of undesirable clutter due to camera perspective and are not as specific as point outputs. Second, our formulation is general enough to enable various robotic tasks. For example, the predicted points can be interpreted as waypoints for navigation, contact points for grasping or region proposals for placement. This not only allows the model to perform multiple tasks but also means it can be trained with multi-task data. 

\begin{figure}
    \centering
    \includegraphics[width=\linewidth]{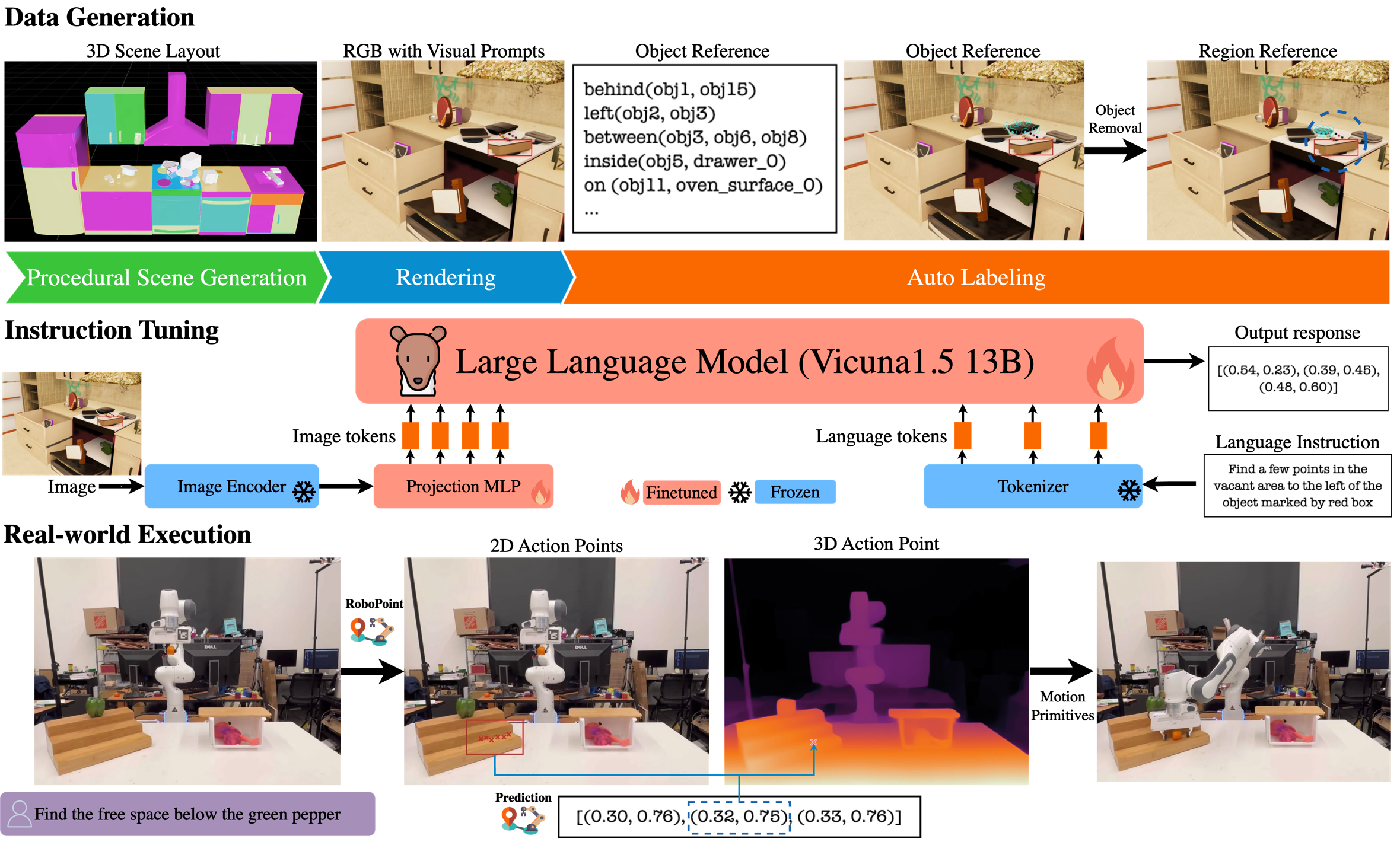}
    \caption{\small \textbf{Overview of \method\ pipeline.} An RGB image is rendered from a procedurally generated 3D scene. We compute spatial relations from the camera's perspective and generate affordances by sampling points within object masks and object-surface intersections. These instruction-point pairs fine-tune the language model. During deployment, \method\ predicts 2D action points from an image and instruction, which are projected into 3D using a depth map. The robot then navigates to these 3D targets with a motion planner.}
    \vspace{-10pt}
    \label{fig:method}
\end{figure}

\paragraph{Instruction Fine-tuning}
\citet{min2022rethinking} has shown that rather than learning new tasks, in-context learning \cite{brown2020language} works by activating patterns from the training data.
Thus, instead of mining patterns from the non-public training dataset, we opt to build our own dataset (see Sec.~\ref{sec:dataset}) and fine-tune the language model's parameters. Specifically, we follow the instruction tuning pipeline in \citet{liu2023llava}. As shown in Fig.~\ref{fig:method}, the model consists of an image encoder, a MLP projector, a language tokenizer and a transformer language model. The image encoder processes the image into a set of tokens which are then projected by a 2-layer MLP into the same space as the language tokens. The multimodal tokens are concatenated and passed through the language transformer. All modules are initialized with pre-trained weights. The projector and the transformer weights are allowed to update while the vision encoder and tokenizer weights are frozen. The model is autoregressive and the objective is to predict the response tokens and a special token delineating the boundary between instruction and response. Our results (Table~\ref{tab:affordance}, Fig.~\ref{fig:real_world}) show that our instruction-tuned model achieves much higher precision than baselines using in-context learning \cite{nasiriany2024pivot,chen2024spatialvlm}.

\paragraph{Co-finetuning with Synthetic Data}
We find that providing the appropriate mix of data is crucial to the model's performance on downstream tasks. As observed by \citet{brohan2023rt}, co-training with a mix of robotic data and internet data ensures the model does not forget the knowledge it has learned during pre-training. Our dataset for fine-tuning consists of 4 different sources, as illustrated in Table.~\ref{tab:data_mix}. The VQA data is a mix of 665K conversations from \cite{liu2023improvedllava} where the model is asked to answer questions in natural language based on the input image. This ensures the model can reason in language. The LVIS data is converted from \cite{gupta2019lvis}, where the model is asked to predict bounding box center and dimensions for all instances of a certain category. This teaches the model how to ground language to image regions. The last two data sources, object reference and free space reference, are from our synthetic data pipeline (Sec.~\ref{sec:dataset}), where the object is to identify points on an object or a vacant region, satisfying certain spatial relations. These data enable the VLM to generate precise action points. We formulate different data sources into the same format and co-train with all of them. Table~\ref{tab:ablation} evaluates the importance of each component in our data mix.

\section{Dataset}
\label{sec:dataset}
We generate a diverse dataset in simulation by procedurally randomizing scene layouts, objects, and camera viewpoints. A novel aspect of our pipeline is generating affordance in free space, allowing the model to detect regions without distinct visual cues.

\paragraph{Procedural Scene Generation in Simulation}
To train \method, we generate a large photorealistic dataset in simulation annotated with affordance points. Most existing robotics datasets \cite{mccormac2017scenenet,savva2019habitat,Xiang_2020_SAPIEN,ehsani2021manipulathor} only have a handful of fixed artist-designed scene layouts which limits the types of relations that can be generated. Several recent works have demonstrated the efficacy of procedural scene generation in improving synthetic data diversity \cite{procthor} and robustness during sim2real transfer for different robotics tasks \cite{murali2023cabinet,fishman2023motion}. 
We create a diverse dataset by procedurally randomizing several aspects of the scene: the 3D layouts, objects and camera view points. The scene is represented as a articulated body, including revolute (e.g. fridge, dishwasher doors) as well as prismatic joints (e.g. cabinet drawers). Objects are sampled from a large repository \cite{eppner2021acronym} with over 8K instances and 262 categories. The objects can be placed on any support surface. This allows our model to learn relations in a truly 3D environment.
Once the 3D scene is created, we compute spatial relations among the objects and render an image for each relation from a diverse set of viewpoints in parallel. 
The diverse view distribution allow \method\ to maintain a consistent prediction across different viewpoints (Fig.~\ref{fig:multiview}). Around 660K (image, relation) pairs are generated from 10K scenes. Some examples from the dataset are shown in Table~\ref{tab:data_mix}. More details can be found in Sec.~\ref{sec:data}.


\begin{table}
    \centering
    \scriptsize
    \begin{tabularx}{\linewidth}{lXXXX}
        \toprule
        Source & Object Reference & Free Space Reference & VQA \cite{liu2023improvedllava} & LVIS \cite{gupta2019lvis} \\ \midrule
        & \includegraphics[width=\linewidth]{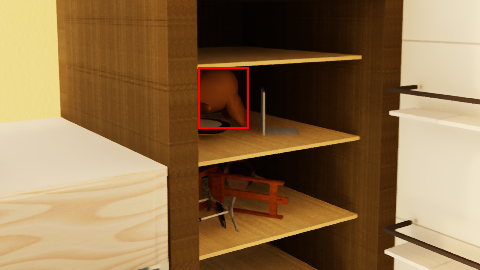} & \includegraphics[width=\linewidth]{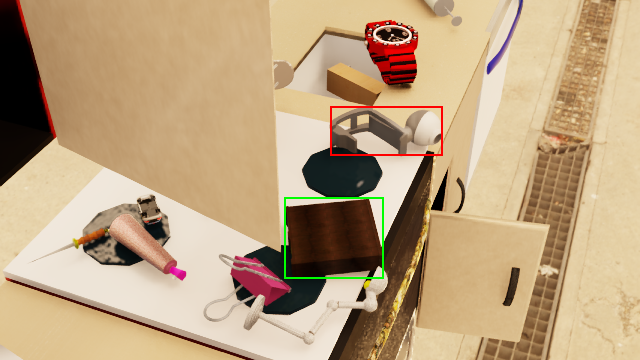} & \includegraphics[width=\linewidth]{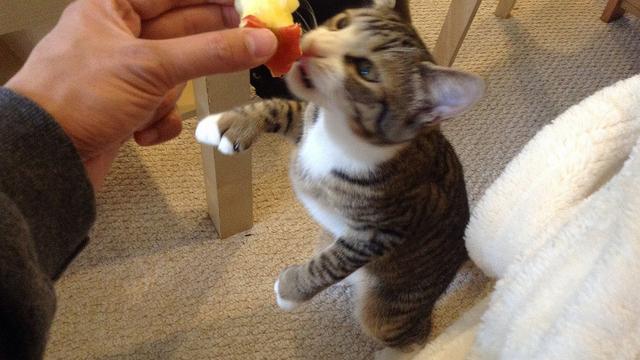} & \includegraphics[width=\linewidth]{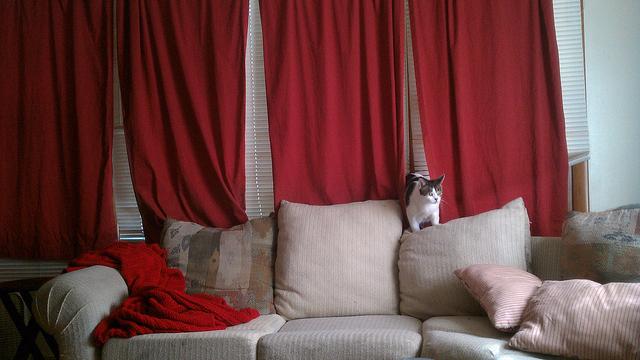} \\ \midrule
        Quantity & 347K & 320K & 665K & 100K \\ \midrule
        Query & 
        Locate several points on an item situated beneath the bordered item. & 
        Find some spots within the vacant space situated between the marked items. & 
        What is the person feeding the cat? & 
        Find all instances of cushions. \\ \midrule
        Answer & [(0.56, 0.69), (0.53, 0.76), (0.45, 0.72), (0.43, 0.67)] 
        & [(0.57, 0.48), (0.58, 0.49), (0.56, 0.45), (0.55, 0.47)] 
        & The person is feeding an apple to the cat. & 
        [(0.49, 0.38, 0.08, 0.06), (0.53, 0.42, 0.07, 0.05)] 
        \\ \bottomrule
    \end{tabularx}
    \vspace{2pt}
    \caption{\small \textbf{Our dataset for instruction-tuning} combines object and space reference data with VQA and object detection data. \method\ leverages spatial reasoning, object detection, and affordance prediction from these diverse sources, enabling it to generalize combinatorially.}
    \vspace{-15pt}
    \label{tab:data_mix}
\end{table}

\paragraph{Generating Affordance in Free Space}
A key novelty in our data pipeline is the generation of affordance in free space. This allows \method\ to detect regions without distinct visual cues, e.g. ``the left part of pizza box" in Fig.~\ref{fig:real_world}, which an off-the-shelf object detector will not be able to detect.
We employ a simple yet effective strategy.
Namely, we first compute relations between a target object and another object or surface. Then, we remove the target object, re-render the image, and sample points inside the intersection of the target object mesh and the surface supporting it. This creates affordance labels in free space in relation to other entities in the scene. 

\section{Experimental Results}
\label{sec:result}
We demonstrate that \method\ achieves superior accuracy in spatial affordance prediction and real-world language-conditioned manipulation than state-of-the-art VLMs \cite{liu2024llavanext,openai2024gpt4o} and visual prompting methods \cite{nasiriany2024pivot,chen2024spatialvlm}. Its view-point consistent prediction and conversational ability also enables application to navigation and augmented reality.


\subsection{Spatial Affordance Prediction}
\method\ significantly outperforms baselines in terms of accuracy on pointing to objects and free space referred by language. In addition, it generalizes to novel relation types, respects physical constraints, maintains common sense knowledge and produces view-consistent predictions.

\paragraph{Benchmarks}
We evaluate spatial affordance prediction on two problems: object reference and free space reference. The object reference data is a 750-image subset of RoboRefIt \cite{lu2023vl}. Unlike human-centered dataset such as RefCoco \cite{yu2016modeling}, RoboRefIt features cluttered images with similar-looking objects that can only be distinguished by relational references.

\begin{figure}[t]
    \centering
    \begin{subfigure}{0.30\linewidth}
        \includegraphics[width=\linewidth]{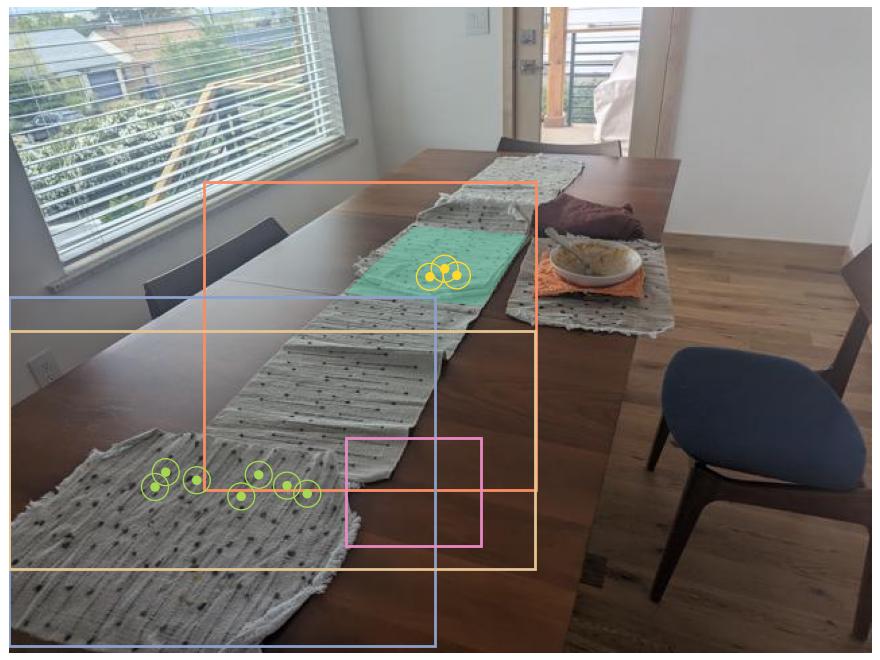}
        \vspace{-15pt}
        \caption{left of bowl and on the tarp}
        \label{fig:combo_relation}
    \end{subfigure}
    \begin{subfigure}{0.30\linewidth}
        \includegraphics[width=\linewidth]{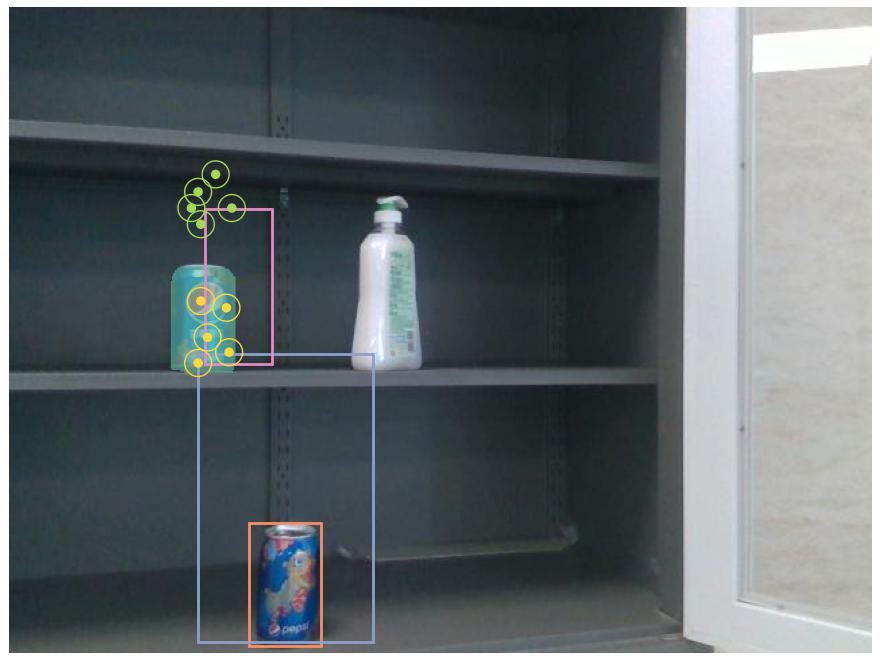}
        \vspace{-15pt}
        \caption{pepsi can on the middle shelf}
        \label{fig:unseen_relation}
    \end{subfigure}
    \begin{subfigure}{0.375\linewidth}
        \includegraphics[width=\linewidth]{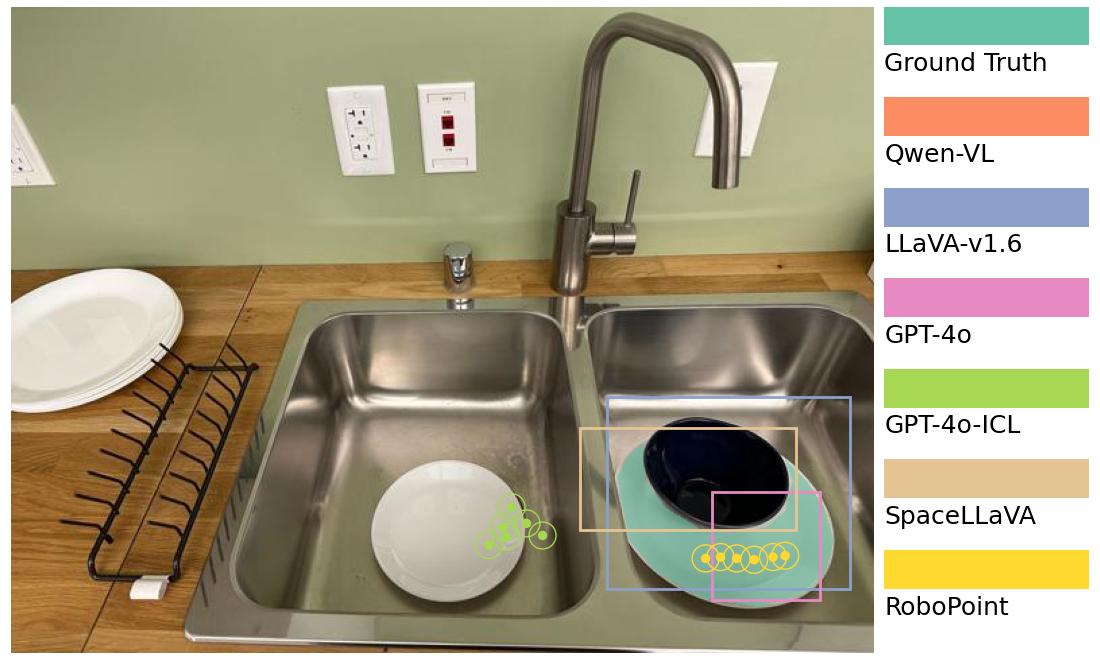}
        \vspace{-15pt}
        \caption{on the rightmost white plate}
        \label{fig:obstacle}
    \end{subfigure}
    \vspace{-5pt}
    \caption{\small \textbf{Visualization of spatial affordance prediction on objects and free space.} \method\ can generalize to (a) combinations of seen relations; (b) unseen relations and (c) scenarios with physical constraints.}
    \label{fig:qualitative}
\end{figure}

Unlike object reference, no existing dataset addresses identifying \textit{free space}. Therefore, we collect \dataset, a dataset of 100 real-world images from homes and offices in the wild. To minimize bias, we ask one group to label each image with an instruction describing a vacant region relative to other entities, and a different group to label masks according to the instruction. As shown in Fig.~\ref{fig:qualitative}, \dataset features diverse and challenging scenes with clutter. A subset of 30 examples (\dataset\ (h)) contain relation types not in our synthetic data.

\paragraph{Baselines}
We compare \method\ against 3 state-of-the-art VLMs, Qwen-VL \cite{bai2023qwen}, LLaVA-NeXT \cite{liu2024llavanext}, GPT-4o \cite{openai2024gpt4o} as well as SpaceLLaVa \cite{remyxai2024spacellava}, a community implementation of SpatialVLM \cite{chen2024spatialvlm}. We employ a zero-shot visual prompting strategy effective for pretrained VLMs. We label the input image with axes indicating its dimensions and ask the model to output a bounding box (top-left and bottom-right corners) of the target object/region, then sample evenly within the bounding box. For GPT-4o, we also tested in-context learning (GPT-4o-ICL) by providing 14 input-output pairs from our synthetic dataset as context before the query. In-context learning achieved zero accuracy for Qwen-VL and LLaVA-Next, likely because point outputs were not part of their training data.

\begin{table}
    \centering
    \scriptsize
    \setlength{\tabcolsep}{3pt}
    \begin{tabular}{ccccccc}
        \toprule & Qwen-VL \cite{bai2023qwen} & LLaVA-NeXT-34B \cite{liu2024llavanext} & SpaceLLaVA \cite{remyxai2024spacellava} & GPT-4o \cite{openai2024gpt4o} & GPT-4o-ICL \cite{openai2024gpt4o} & \method\ \\
        \toprule
        RoboRefIt \cite{lu2023vl} & $24.08\pm0.85$ & $19.91\pm0.92$ & $21.30\pm0.87$ & $15.28\pm1.27$ &$ 9.01\pm6.45$ & $\bf 49.82\pm0.52$ \\
        \dataset\ & $10.49\pm0.77$ & $15.02\pm0.88$ & $11.84\pm0.73$ & $29.06\pm1.33$ & $14.46\pm6.38$ & $\bf 46.77\pm0.45$ \\
        \dataset\ (h) & $9.90\pm0.22$ & $14.76\pm2.42$ & $12.10\pm1.36$ & $27.14\pm1.47$ & $14.83\pm4.68$ & $\bf 44.48\pm1.35$ \\ \bottomrule
    \end{tabular}
    \vspace{3pt}
    \caption{\textbf{Quantitative comparisons on object reference (RoboRefIt) and free space reference (\dataset).} \method\ outperforms state-of-the-art VLMs by a significant margin, even on examples where the spatial relations are unseen during fine-tuning (\dataset\ (h)). The metric is percentage of predicted points within the target mask.}
    \label{tab:affordance}
\end{table}

\begin{table}[h!]
    \centering
    \vspace{-1em}
    \scriptsize
    \setlength{\tabcolsep}{4pt}
    \begin{tabular}{cccccccccc}
        \toprule & GQA \cite{hudson2019gqa} & MME \cite{fu2023mme} & POPE \cite{li2023evaluating} & RefCoco \cite{yu2016modeling} & SEED \cite{li2023seed} & TextVQA \cite{singh2019towards} & VizWiz \cite{gurari2018vizwiz} & VQA-v2 \cite{goyal2017making} \\ \midrule
        LLaVA-13B \cite{liu2023improvedllava} & 63.24 & 1522.59 & 85.92 & 31.99 & 67.06 & \textbf{48.73} & 56.65 & \textbf{78.26} \\
        \method\ & \textbf{63.28} & \textbf{1524.78} & \textbf{86.01} & \textbf{32.16} & \textbf{67.52} & 47.31 & \textbf{60.37} & 77.83 \\ \bottomrule
    \end{tabular}
    \vspace{3pt}
    \caption{\textbf{Quantitative evaluation on standard VQA benchmarks.} \method\ performs on par with state-of-the-art VLM, maintaining the common sense knowledge learned from pretraining.}
    \label{tab:vqa}
\end{table}

\begin{table}[h!]
    \centering
    \vspace{-1em}
    \small
    \begin{tabular}{cccccc}
        \toprule
        No VQA \cite{liu2023improvedllava} & No LVIS \cite{gupta2019lvis} & No Object Ref & No Space Ref & 10\% Data & All \\ \midrule
        $28.28\pm2.08$ & $34.27\pm0.62$ & $42.23\pm2.28$ & $13.21\pm1.04$ & $15.71\pm0.77$ & $\bf 46.77\pm0.45$ \\ \bottomrule
    \end{tabular}
    \vspace{3pt}
    \caption{\textbf{Ablation on the data composition.} Results on \dataset show that best results are achieved when all of the data sources are combined during instruction-tuning.}
    \label{tab:ablation}
    \vspace{-1em}
\end{table}

\paragraph{Results}
In Table~\ref{tab:affordance}, we report the average prediction accuracy for \method\ and the baselines along with standard deviation computed from 3 different runs. The accuracy is calculated as the percentage of predicted points within the ground truth target mask. We can see that \method\ achieves significantly higher accuracy than all baselines, demonstrating the power of \method\ in spatial reasoning and precise target generation. Some results are visualized in Fig.~\ref{fig:qualitative}.

\begin{figure}
    \centering
    \vspace{-1em}
    \includegraphics[width=0.28\linewidth]{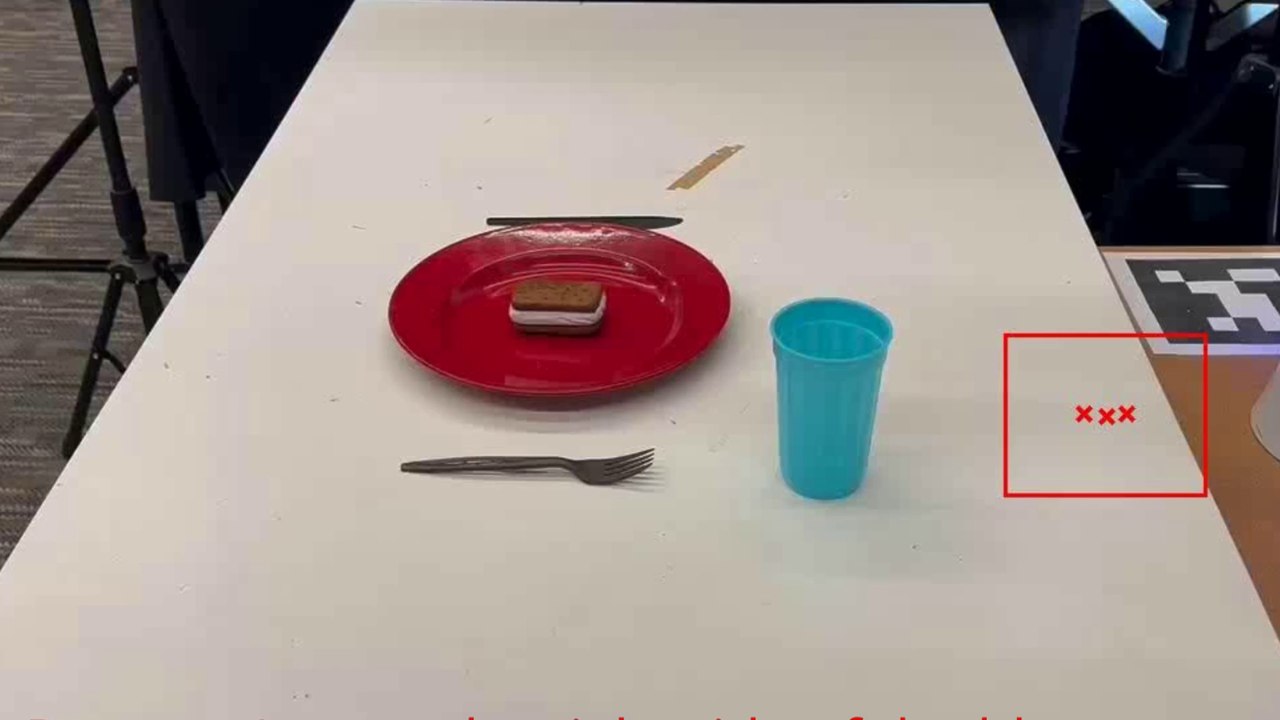}
    \includegraphics[width=0.28\linewidth]{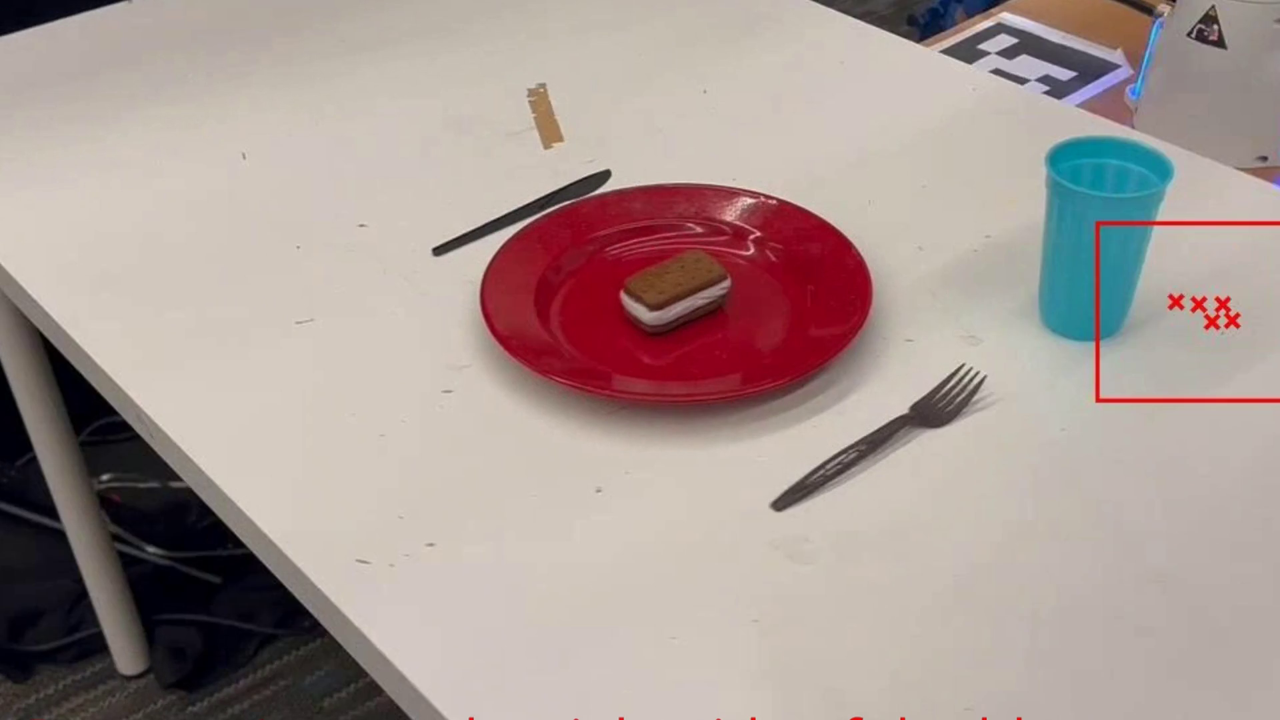}
    \includegraphics[width=0.28\linewidth]{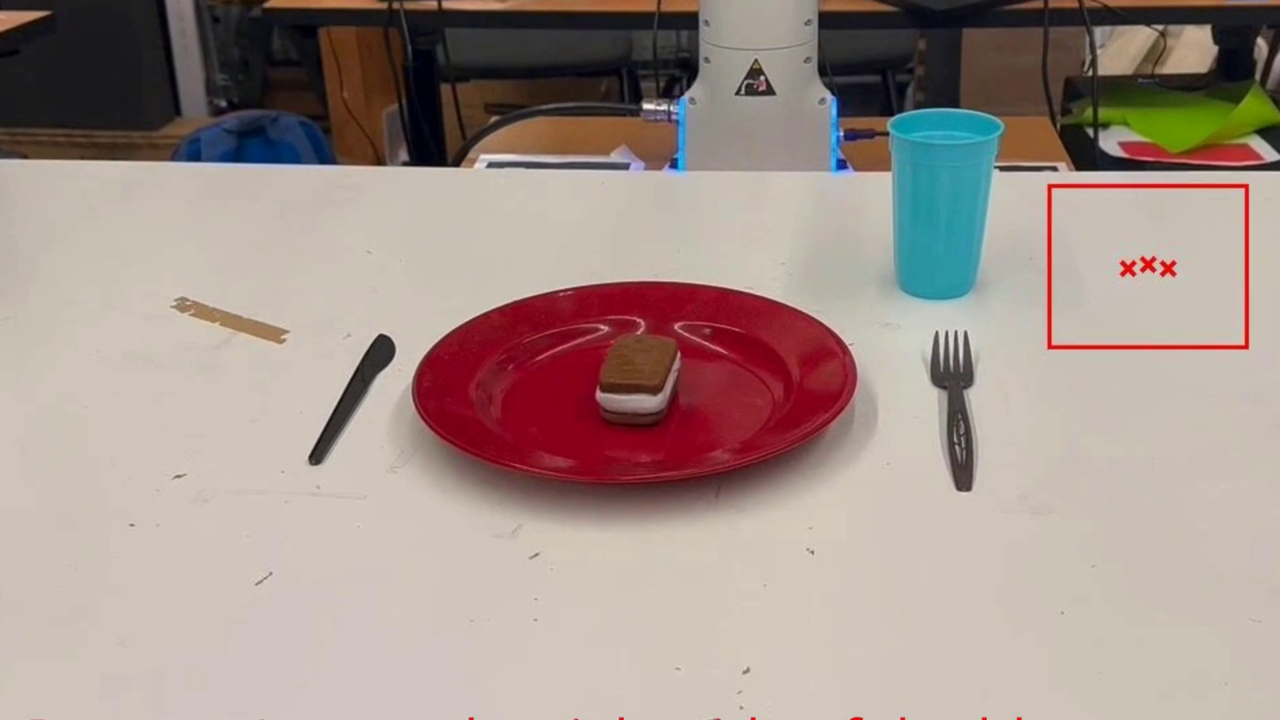}
    \caption{\textbf{\method's prediction is consistent across different viewpoints.} Red cross shows \method's response to ``find free space right of the blue cup" in different views.}
    \label{fig:multiview}
\end{figure}

\begin{figure}
  \centering
  \vspace{-15pt}
  \includegraphics[width=0.9\linewidth]{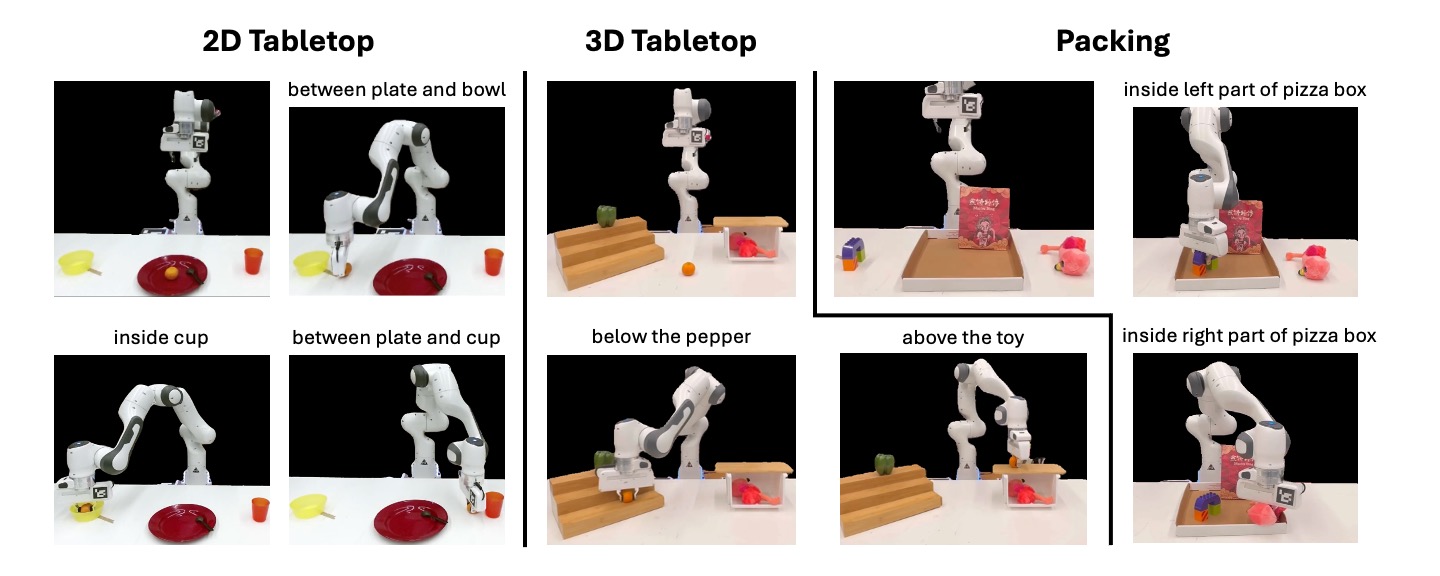}
  {\scriptsize
  \begin{tabular}{cccccccc}
      \toprule & inside cup & plate $\leftrightarrow$ bowl & plate $\leftrightarrow$ cup & below pepper & above toy & inside left part & inside right part \\ \midrule
      Qwen-VL \cite{bai2023qwen} & 3 / 10 & 2 / 10 & 1 / 10 & 0 / 10 & 1 / 10 & 2 / 10 & 2 / 10 \\
      GPT-4V \cite{achiam2023gpt} & 4 / 10 & 1 / 10 & 1 / 10 & 4 / 10 & 2 / 10 & 1 / 10 & 0 / 10 \\
      PIVOT \cite{nasiriany2024pivot} & 1 / 10 & 1 / 10 & 0 / 10 & 2 / 10 & 2 / 10 & 1 / 10 & 0 / 10 \\
      \method\ & \textbf{6} / 10 & \textbf{7} / 10 & \textbf{5} / 10 & \textbf{8} / 10 & \textbf{5} / 10 & \textbf{4} / 10 & \textbf{6} / 10 \\ \bottomrule
  \end{tabular}}
  \caption{\small \textbf{Real-world manipulation evaluation.} We created 7 language-conditioned manipulation tasks to measure \method's capability on real robot. \method outperforms the best baseline by 39.5\% on average success rate, which depends critically on the alignment between the point predictions and the language.}
  \label{fig:real_world}
  \vspace{-1em}
\end{figure}

\paragraph{Does \method\ generalize to unseen relation types?}
The synthetic dataset we constructed in Sec.~\ref{sec:dataset} contains templated language and a fixed set of relations. Nevertheless, \method\ is able to produce accurate predictions for combinations of seen relations (Fig.~\ref{fig:combo_relation}) and novel relation types such as in the middle, rightmost etc. that are not in the fine-tuning dataset (Fig.~\ref{fig:unseen_relation}). It is also able to maintain its advantage over baselines on these novel relations (Table~\ref{tab:affordance}).

\paragraph{Does \method\ respect physical constraints?}
\method's outputs not only satisfy the spatial relations but also respect physical constraints. The target points generated by \method\ avoid obstacles such as the the bowl in Fig.~\ref{fig:obstacle}, whereas the baselines fail to do so.

\paragraph{Does \method\ keep common sense knowledge?}
We evaluate \method's performance on VQA benchmarks and summarize the results in Table~\ref{tab:vqa}. \method\ performs on-par with LLaVA-v1.5-13B \cite{liu2023improvedllava}, a VLM trained on the same pre-trained weights as \method\ on VQA data. This shown that \method\ serves a generic VLM rather than a domain-specific model.

\paragraph{How important is each component in the data mix?}
In Table~\ref{tab:ablation}, we evaluated the importance of each data component on the \dataset\ benchmark. Each data component -- VQA on real images, object detection from LVIS, object and free space reference on synthetic images -- significantly contributes to overall accuracy. This highlights the value of a general problem formulation that incorporates diverse data sources. Additionally, data quantity is crucial, as the model's performance drops significantly when fine-tuned on only 10\% of the data.

\paragraph{Are RoboPoint's predictions consistent across views?}
As shown in Fig.~\ref{fig:multiview}, \method\ maintains consistent predictions with camera movement. This makes it particularly suitable for mobile platforms and AR, where \method\ provides consistent action suggestions with moving cameras. Videos can be found on the project page \href{https://robo-point.github.io/}{robo-point.github.io}.

\subsection{Downstream Applications}
To assess \method's capabilities on downstream robotics and vision tasks, we curated various scenarios for manipulation, navigation and AR assistance. We demonstrate \method's superior performance against state-of-the-art baselines on these tasks. Recordings of robot executions can be found on the project page \href{https://robo-point.github.io/}{robo-point.github.io}.

\paragraph{Real-World Manipulation}
We set up 3 manipulation environments with 7 tasks (Fig.~\ref{fig:real_world}). The robot processes image observations and language commands through \method, which returns 2D point targets. These targets are converted to 3D points using a depth map (Fig.~\ref{fig:method}). The robot's end-effector pose is computed from these 3D points plus an offset. A motion planner then executes the trajectory to the target pose. Success is determined by collision-free execution and accurate placement of the target object as per the language instruction. We conducted 10 trials per task and compared \method\ against zero-shot VLMs like Qwen-VL \cite{bai2023qwen} and GPT-4V \cite{achiam2023gpt}, as well as iterative prompting methods such as PIVOT \cite{nasiriany2024pivot}. \method\ surpasses GPT-4V, the best-performing baseline, by a margin of 39.5\% on average success rate. It also enables new capabilities. 
For instance, in the packing scene, \method's relational reasoning allowed the robot to differentiate regions within a pizza box, fitting multiple objects accurately.

\begin{figure}
  \centering
  \vspace{-1em}
  \includegraphics[width=0.85\linewidth]{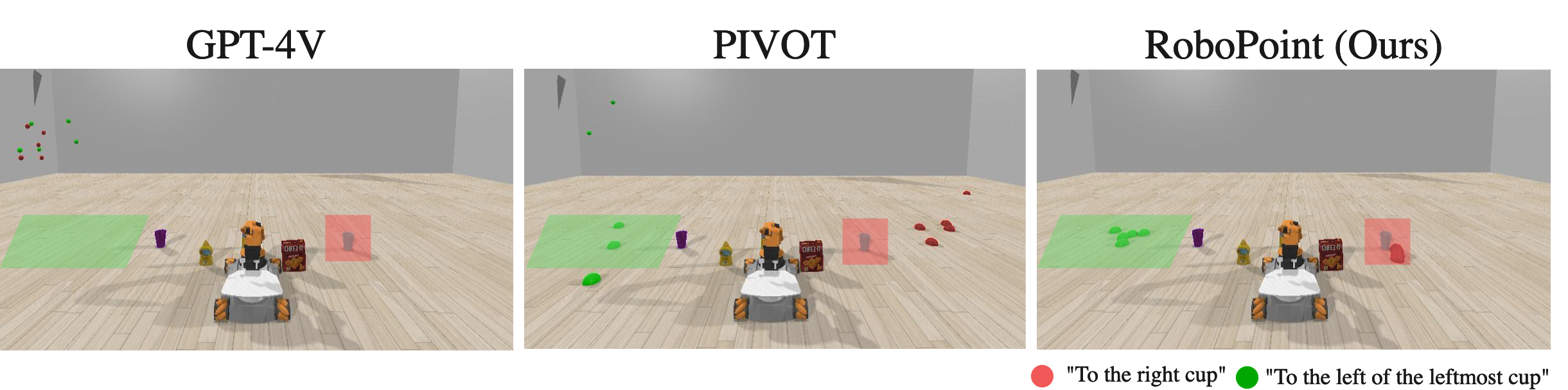}
  {\scriptsize
  \begin{tabular}{cccccccc}
      \toprule & left grill & front grill & right grill & left of left cup & right cup & fridge $\leftrightarrow$ oven & oven $\leftrightarrow$ drawer \\ \midrule
      GPT-4V \cite{achiam2023gpt} & 0 / 5 & 0 / 5 & 0 / 5 & 0 / 5  & 0 / 5 & 0 / 5  & 0 / 5 \\
      PIVOT \cite{nasiriany2024pivot} & 2 / 5 & 3 / 5 & 2 / 5 & 3 / 5 & 4 / 5 & \textbf{3} / 5 & \textbf{2} / 5 \\
      \method\ & \textbf{5} / 5 & \textbf{5} / 5 & \textbf{5} / 5 & \textbf{5} / 5 & \textbf{5} / 5 & 1 / 5 & 1 / 5 \\ \bottomrule
  \end{tabular}}
  \caption{\small \textbf{Application to navigation.} \method\ predicts accurate goal point based on language, leading to higher target reaching rate than GPT-4V and PIVOT. Ground truths are drawn as colored masks and predictions are drawn as colored spheres.}
  \label{fig:sim_nav}
\end{figure}

\begin{figure}
    \centering
    \vspace{-1em}
    \includegraphics[width=0.9\linewidth]{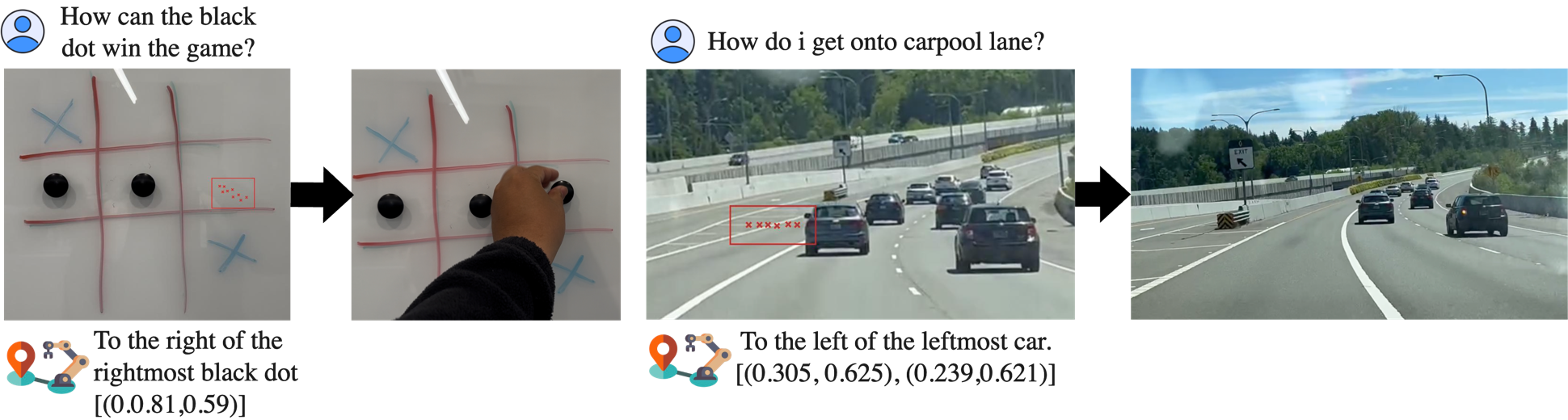}
    \caption{\small \textbf{Application to Augmented Reality.} Given a user query, \method\ first generates natural language response using common sense and then provide visual guidance using spatial affordance prediction, which the user can execute with greater ease than language guidance. }
    \label{fig:AR}
    \vspace{-1em}
\end{figure}




\paragraph{Navigation}
To evaluate \method's spatial affordance predictions beyond tabletop scenarios, we created 3 room scenes using the YouBot mobile manipulation platform in CoppeliaSim \cite{coppeliaSim}, where the robot is tasked to navigate to a target region with respect to certain entities in the scene. Fig.~\ref{fig:sim_nav} shows the distribution of affordance generated by \method, PIVOT \cite{nasiriany2024pivot} and GPT-4V \cite{achiam2023gpt} and the success rate of navigating to the correct region using the predicted points with a simple path planner. \method\ outperforms PIVOT and GPT-4V in 2 out of 3 scenarios, demonstrating its effectiveness in large-scale room environments for navigation.

\paragraph{Augmented Reality} 
\method, which is co-trained with VQA data, retains conversational capabilities in natural language. As demonstrated in Fig.~\ref{fig:teaser}, users can interact with \method\ through language and receive action suggestions visually with the predicted affordance. In addition to the \textit{set a formal dining table} task in Fig.~\ref{fig:teaser}. We demonstrate two more real-world scenarios-\textit{win tic-tac-toe} and \textit{get to carpool lane}-in Fig.~\ref{fig:AR}, where \method\ gives visual guidance to solve the tasks by predicting the correct spatial affordance points.

\section{Conclusion}
\label{sec:conclusion}
We propose \method, a novel VLM designed to predict spatial affordances in images based on relational language instructions. By integrating real-world VQA data with automatically generated synthetic data, \method\ is able to generate precise action points that adhere to spatial and physical constraints, overcoming the limitations of current VLMs in robotics, which often rely on pre-defined motion primitives or large-scale expert demonstrations. Experimental results show \method's superior performance in complex tasks, such as relational free space reference and object rearrangement in cluttered environments, compared to state-of-the-art methods. Additionally, \method's versatility extends its applicability to augmented reality and robot navigation, showcasing its potential for broader applications in robotics.

\paragraph{Limitation:} \method does not provide confidence estimates for the point predictions. The number of output points are also not controllable. Both of these are valuable directions to explore in future work.

\clearpage
\acknowledgments{We thank Yi Ru Wang for providing language annotations, Tucker Hermans, Ajay Mandlekar, Jonathan Tremblay, Wei Yang, Jie Xu for providing images in the \dataset dataset.}
\bibliography{references}

\clearpage
\section*{Appendix}
\setcounter{table}{0}
\renewcommand{\thetable}{\Alph{table}}
\setcounter{figure}{0}
\renewcommand{\thefigure}{\Alph{figure}}
\setcounter{section}{0}
\renewcommand{\thesection}{\Alph{section}}

\section{Instruction Tuning} \label{sec:tuning}
\method is instruction-tuned from a Vicuna-v1.5-13B base model \cite{vicuna2023} with a ViT-L/14 336px image encoder pretrained with CLIP \cite{radford2021learning}. The projector is a 2-layer MLP pretrained on the 558K subset of the LAION-CC-SBU dataset with BLIP captions from \cite{liu2023llava}. The instruction tuning took 40 hours on 16 A-100 GPUs with a batch size of 16 per-GPU. The learning rate is set to 4e-5.

\section{Data Generation} \label{sec:data}
Table~\ref{tab:dataset} shows more examples from our procedually generated synthetic dataset for object reference and free space reference.
 
We sample assets that one can find in an typical kitchen environments (e.g. dishwasher, hood, table, fridge) and use heuristics to place them in random, but semantic layouts in the scene. Once the furniture assets are added to the scene. We used a large object dataset sampled from ACRONYM \cite{eppner2021acronym}. Object positions are randomly sampled on support surfaces (e.g. countertop, table) and the orientations are determined by their stable poses. Poses that result in the object being in collision with the existing scene are rejected. We place cameras randomly in the scene and select those with at least three visible objects (visible means the number of points within segmentation mask is larger than 100) and at least 1 valid relationship between a pair of visible objects. 
The diverse view distribution allow \method\ to maintain a consistent prediction across different viewpoints. Around 660K (image, relation) pairs are generated from 10K scenes.

We use the 3D bounding boxes of objects, surfaces and containers in the scene layout to compute a set of pairwise relations, including left, right, in front, behind, above, below, next to, on, inside, on left part, on right part, on front part, on back part. Note that although these relations are templated, the model fine-tuned on these data is able to generalize to new types of relations, as shown in Fig.~\ref{fig:qualitative}. For each relation, we first sample points on the object being referenced to create an example for object reference. Around 1 to 50 ground truth points are sampled per image. We convert the sampled points to a list of image coordinates normalized between 0 and 1 and use that as the ground truth response.

One caveat for these procedurally generated scenes is that the objects do not have rich text descriptions. Most objects just have a category name. We get around this problem by adding visual prompts to the rendered images. Specifically, we draw colored bounding boxes around the objects referenced in the language instruction. As a result, a typical instruction in the synthetic data will look like: ``There is an object surrounded by a red rectangle in the image. Find some places in the free area to the left of the marked object." Note that we do not add these visual prompts during testing, and thus do not require object detection. The idea is that the model learns to detect objects from other sources of data (e.g. LVIS~\cite{gupta2019lvis}), and it will focus on relational reasoning when dealing with the object and space reference data.

\begin{table}
    \centering
    \begin{tabularx}{\linewidth}{lXX}
        \toprule
        Relation & Above & Behind \\ \midrule
        & \includegraphics[width=\linewidth]{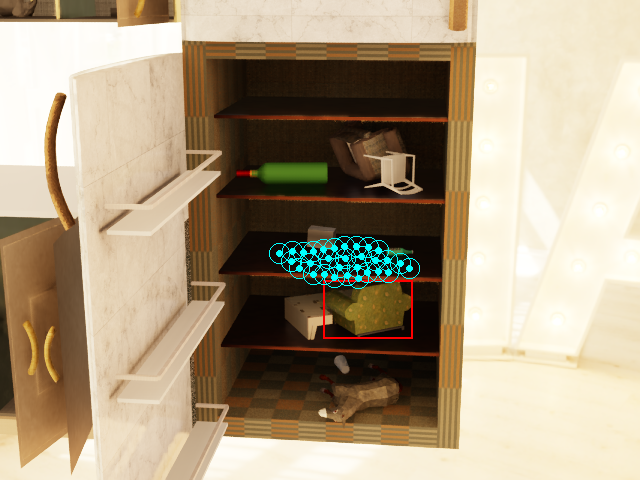} & \includegraphics[width=\linewidth]{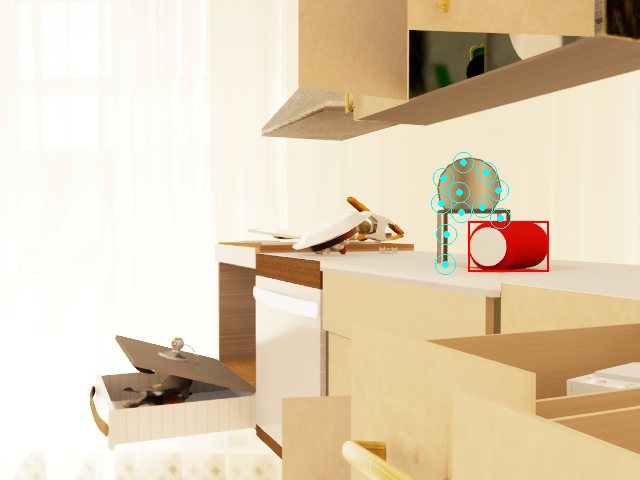} \\ \midrule
        Prompt & The image features an item encased in a red rectangular border. Locate several spots within the vacant space situated above the bordered item. & In the image, an object is framed by a red rectangle. Locate a few points on an object that is situated behind the framed object. \\ \midrule
        Relation & Between & Inside \\ \midrule
        & \includegraphics[width=\linewidth]{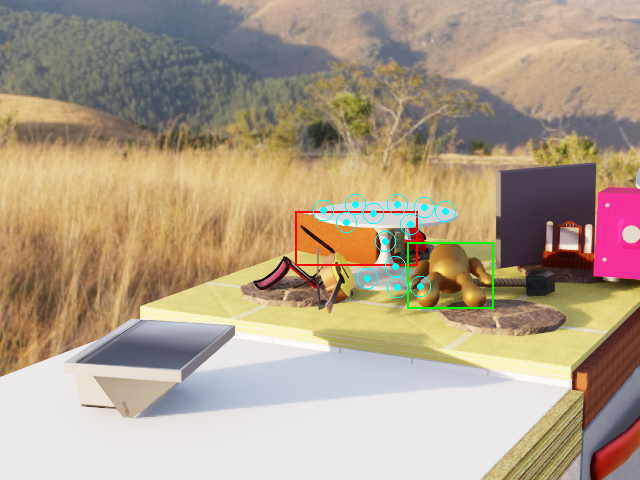} & \includegraphics[width=\linewidth]{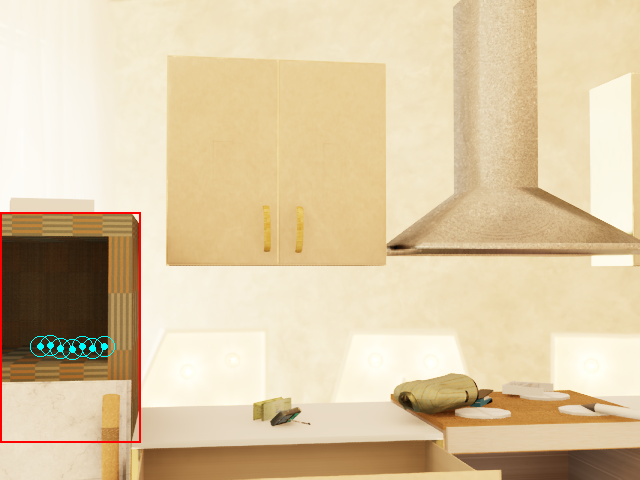} \\ \midrule
        Prompt & In the image, there is an item framed by a red rectangle and another item encased within a green rectangle. Locate several points upon the item situated between the two highlighted items. & The image depicts a container delineated by a red rectangular border. Pinpoint several spots within the vacant area enclosed by the outlined container. \\ \midrule
        Relation & Right & On left part \\ \midrule
        & \includegraphics[width=\linewidth]{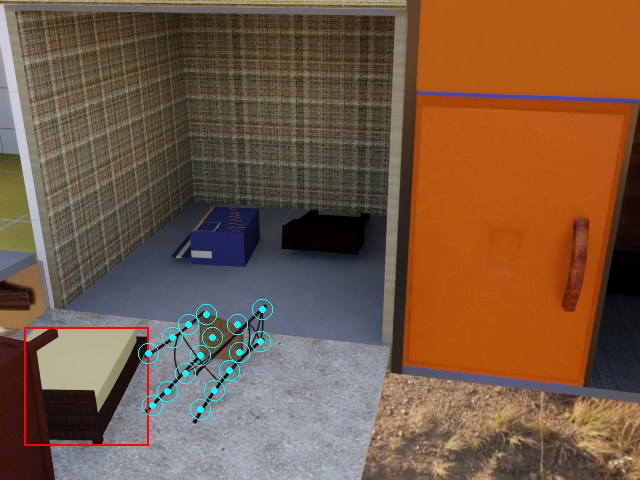} & \includegraphics[width=\linewidth]{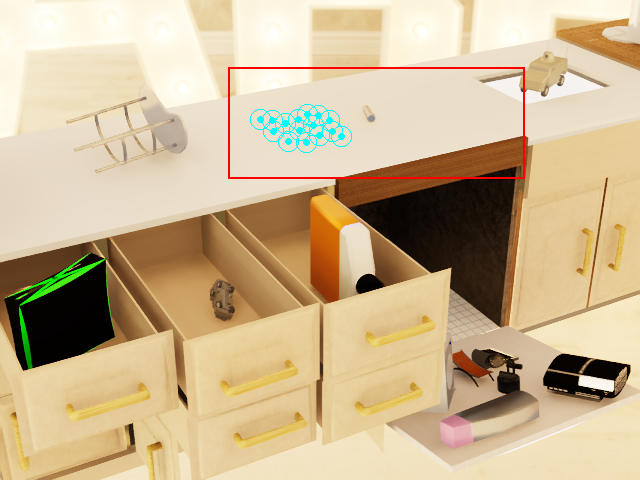} \\ \midrule
        Prompt & The image features an object outlined by a red rectangle. Locate several points on an item that is situated on the right side of the marked item. & The image showcases an area demarcated by a red rectangle. Locate a few points within a vacant area on the right side of the marked surface. \\
        \bottomrule
    \end{tabularx}
    \vspace{5pt}
    \caption{Examples from the synthetic dataset used to teach \method relational object reference and free space reference. The red and ground boxes are visual prompts to indicate reference objects and the cyan dots are the visualized ground truth (not included in the image inputs to the model).}
    \label{tab:dataset}
\end{table}

\section{Qualitative Examples} \label{sec:more_qualitative}
\begin{figure}
    \centering
    \begin{subfigure}{0.44\linewidth}
        \includegraphics[width=\linewidth]{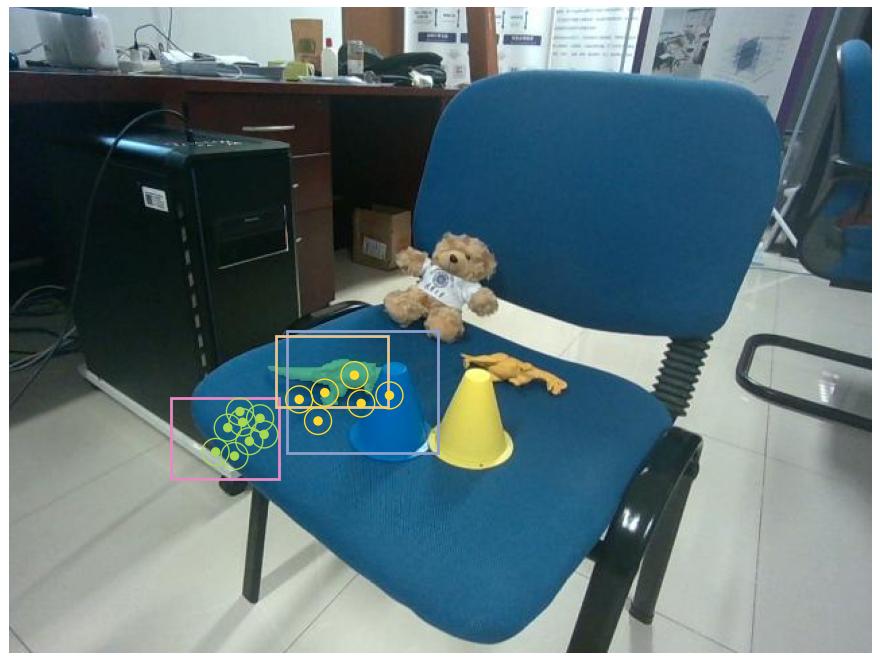}
        \vspace{-15pt}
        \caption{dinosaur model on the left}
        \vspace{5pt}
    \end{subfigure}
    \begin{subfigure}{0.55\linewidth}
        \includegraphics[width=\linewidth]{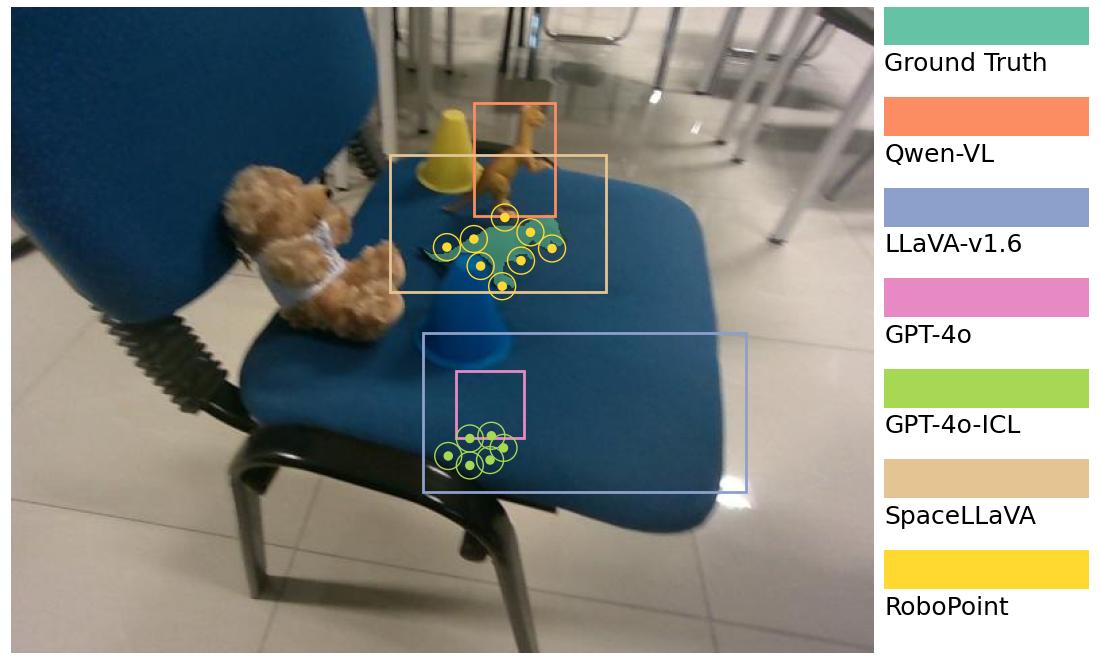}
        \vspace{-15pt}
        \caption{dinosaur model at the bottom}
        \vspace{5pt}
    \end{subfigure}
    \begin{subfigure}{0.44\linewidth}
        \includegraphics[width=\linewidth]{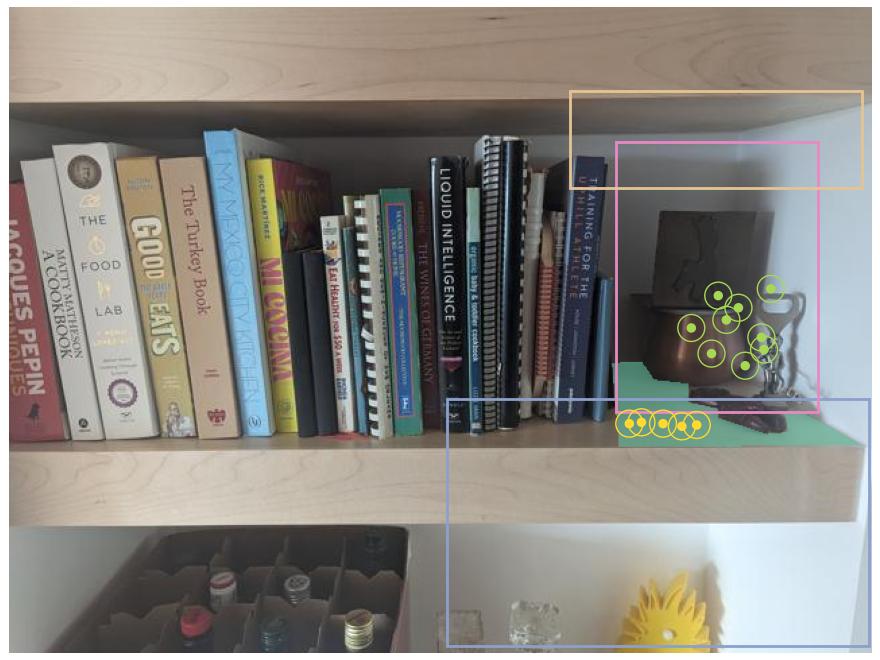}
        \vspace{-15pt}
        \caption{to the right of the books}
        \vspace{5pt}
    \end{subfigure}
    \begin{subfigure}{0.55\linewidth}
        \includegraphics[width=\linewidth]{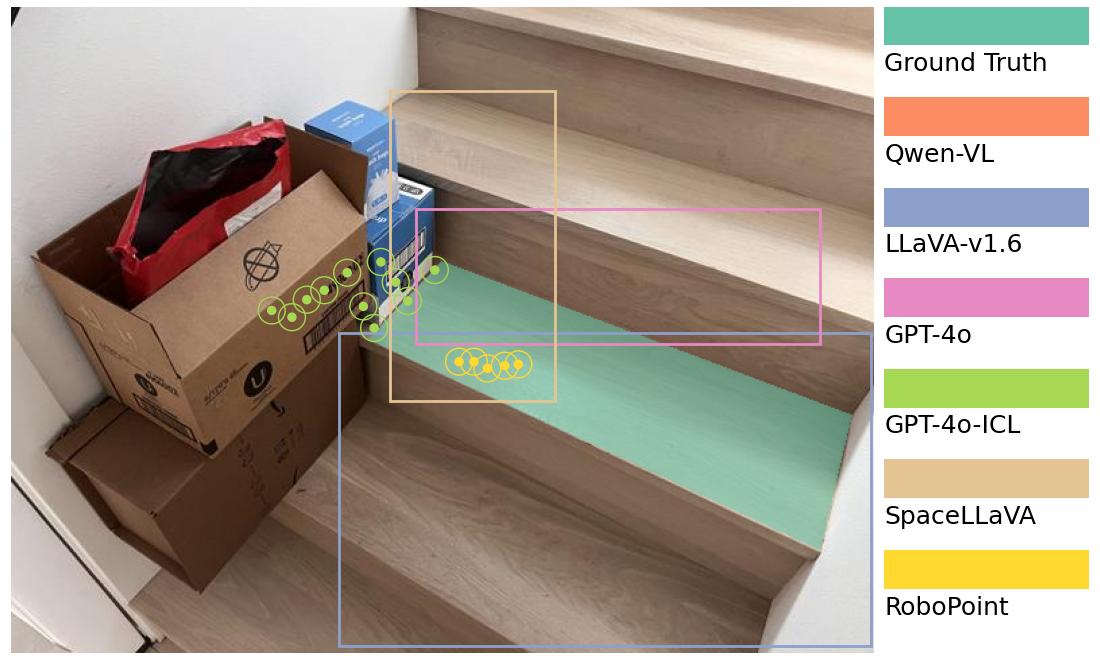}
        \vspace{-15pt}
        \caption{on the stair in the middle}
        \vspace{5pt}
    \end{subfigure}
    \begin{subfigure}{0.44\linewidth}
        \includegraphics[width=\linewidth]{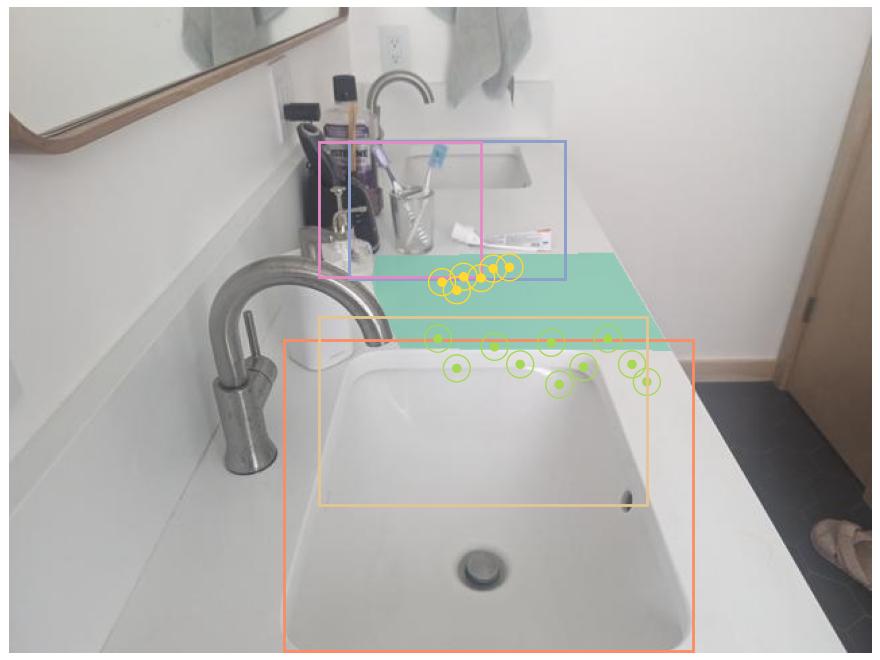}
        \vspace{-15pt}
        \caption{to the rear of the sink in the front}
        \vspace{5pt}
    \end{subfigure}
    \begin{subfigure}{0.55\linewidth}
        \includegraphics[width=\linewidth]{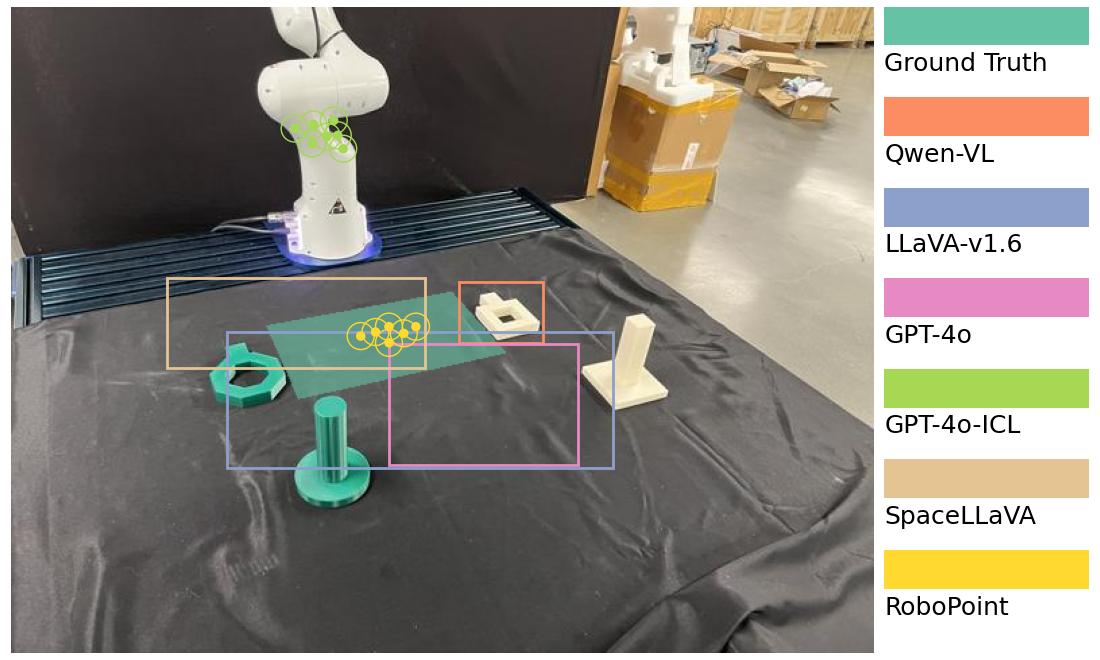}
        \vspace{-15pt}
        \caption{between the green block and the white block in the back}
        \vspace{5pt}
    \end{subfigure}
    \begin{subfigure}{0.44\linewidth}
        \includegraphics[width=\linewidth]{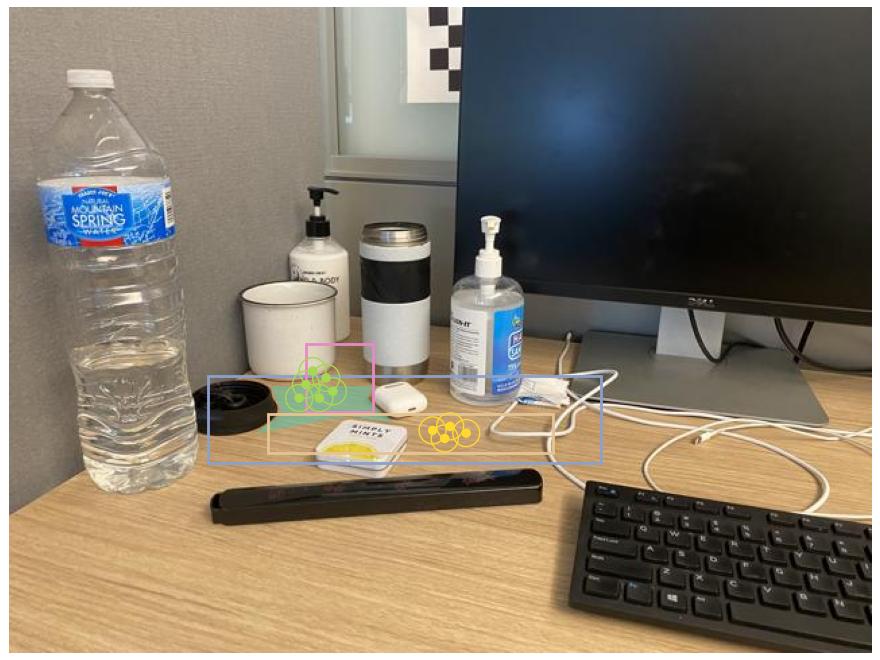}
        \vspace{-15pt}
        \caption{in between the airpods and the black lid}
        \vspace{5pt}
    \end{subfigure}
    \begin{subfigure}{0.55\linewidth}
        \includegraphics[width=\linewidth]{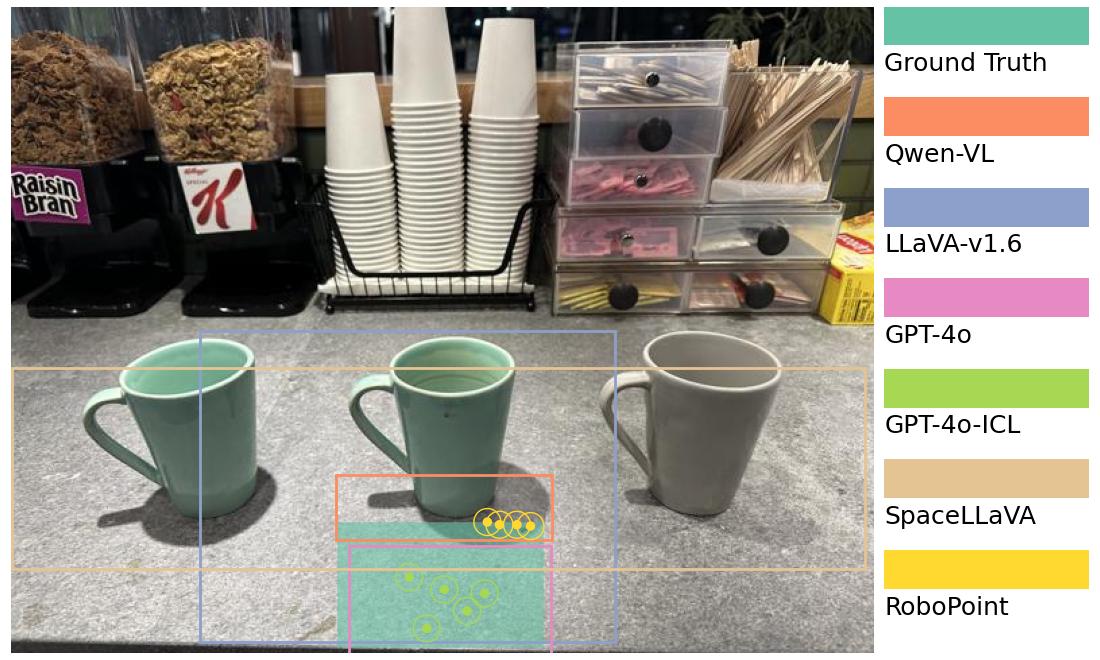}
        \vspace{-15pt}
        \caption{in front of the mug in the middle}
        \vspace{5pt}
    \end{subfigure}
    \caption{Qualitative results on RoboRefIt (a, b) and \dataset (c, d, e, f, g, h), including cases with relations unseen during training (d, e, f, h) and where GPT-4o performs better (g, h).}
    \label{fig:more_qualitative}
\end{figure}

Fig.~\ref{fig:more_qualitative} shows more qualitative comparisons of \method against baselines on RoboRefIt~\cite{lu2023vl} and \dataset data, including examples demonstrating generalization to novel relation types and cases where \method underperforms GPT-4o~\cite{openai2024gpt4o}.

\end{document}